\documentclass[runningheads]{llncs}
\usepackage{graphicx}
\usepackage{booktabs}
\usepackage{amssymb}
\usepackage{subfigure}
\usepackage{amsmath, amssymb}
\usepackage{setspace}
\usepackage{cite}
\usepackage{tabularx}
\usepackage{booktabs}
\usepackage{multirow}
\usepackage{algorithm}
\usepackage{algpseudocode}
\usepackage{hyperref}
\usepackage{marvosym}

\begin{document}
\title{FGeo-DRL:Deductive Reasoning for Geometric Problems through Deep Reinforcement Learning}
\titlerunning{GPS through Deep Reinforcement Learning}

\author{
    Jia Zou\inst{1,2} \and
    Xiaokai Zhang\inst{1} \and
    Yiming He\inst{1,2} \and
    Na Zhu\inst{1,2} \and
    Tuo Leng\inst{1,2}\textsuperscript{(\Letter)}
}

\authorrunning{J. Zou et al.}

\institute{
    School of Computer Engineering and Science, Shanghai University, Shanghai, China \\ \email{tleng@shu.edu.cn} \and
    Institute of Artificial Intelligence, Shanghai University, Shanghai, China
}

\maketitle

\begin{abstract}
The human-like automatic deductive reasoning has always been one of the most challenging open problems in the interdiscipline of mathematics and artificial intelligence. This paper is the third in a series of our works. We built a neural-symbolic system, called FGeoDRL, to automatically perform human-like geometric deductive reasoning. The neural part is an AI agent based on reinforcement learning, capable of autonomously learning problem-solving methods from the feedback of a formalized environment, without the need for human supervision. It leverages a pre-trained natural language model to establish a policy network for theorem selection and employ Monte Carlo Tree Search for heuristic exploration. The symbolic part is a reinforcement learning environment based on geometry formalization theory and FormalGeo, which models GPS as a Markov Decision Process. In this formal symbolic system, the known conditions and objectives of the problem form the state space, while the set of theorems forms the action space. Leveraging FGeoDRL, we have achieved readable and verifiable automated solutions to geometric problems. Experiments conducted on the formalgeo7k dataset have achieved a problem-solving success rate of 86.40\%. The project is available at \href{https://github.com/PersonNoName/FGeoDRL}{here}.

\end{abstract}

\section{Introduction}
The rapid development of large language models signifies a new potential for machine intelligence in addressing geometric problems. The emergence of ChatGPT has significantly enhanced the ability to effectively solve geometric problems by enabling machines to delve deeper into the understanding of human language\cite{chatgpt}. This innovation has paved the way for the application of artificial intelligence in the field of mathematics. Through the language model's comprehension and analysis of geometric problems\cite{inter-gps}, users can obtain more intuitive and insightful explanations, along with targeted suggestions, such as selecting appropriate geometric principles or proof methods.

However, it is important to note that, despite the positive role played by large language models in the initial resolution of geometric problems, they are not mathematical experts. The advice they provide should be regarded as inspirational guidance rather than absolute and precise theorem proofs\cite{G-LLaVA}. When dealing with more complex and formal geometric proofs, reliance on professional mathematical knowledge and tools, such as mathematical software and theorem provers, is still necessary.

Nevertheless, relying solely on language models for geometric reasoning raises doubts in both the interpretability of theorem application and its executability\cite{Emergent}. We are uncertain about how neural networks consider the information provided in our input problems, and when it comes to the output's executable sequences, accurate measurement is lacking to prove the fulfillment of solving the problem entirely through the execution of the output theorem sequence. Logical natural language with step-by-step derivation depends on human discernment, making it impossible to directly infer the solvability of geometric problems through natural language model reasoning\cite{Sequence, Sequence2}.

To address the aforementioned issues, we propose a geometric formal language system called FormalGeo. By transforming natural language and graphical information into a universally machine-processable formal language, we have successfully integrated multiple existing geometric datasets, including Geometry3k\cite{inter-gps}, GeoQA\cite{GeoQA}, GeoQA+\cite{GeoQA+}, and online resources. To enhance data quality, we conducted processes such as manual annotation, deduplication, and correction, comprehensively optimizing the dataset. Ultimately, we successfully annotated approximately 6981 problems, creating a novel geometric dataset named FormalGeo7k.

Due to the extensive need for annotated datasets in training natural language models, the limited dataset annotated through our formal language system is insufficient for complete theorem prediction using natural language models. We propose a geometric problem-solving approach based on reinforcement learning, named FormalGeo Deep Reinforcement Learning (FGeoDRL). We address the aforementioned challenges by introducing reinforcement learning for search, continually interacting with the formalized system environment to generate new data and adapt learning strategies for theorem prediction. Simultaneously, reinforcement learning often requires defining a suitable reward function for guidance. In the case of immediate rewards, manually defining them is subjective and faces challenges in dealing with almost infinite state spaces. On the other hand, learning rewards through inverse reinforcement learning requires extensive annotated data and presents interpretability issues\cite{Rewards, Rewards2}. We adopt delayed rewards to handle the reward feedback generated during the search. Utilizing Monte Carlo Tree Search allows us to obtain a search trajectory chain, and the reward value is fed back based on whether the problem is ultimately solved. The search process is guided by the magnitude of rewards generated through multiple simulations\cite{AlphaGo}.

In summary, our contributions can be categorized into the following three points:
\begin{enumerate}
    \item We integrated existing well-known datasets and added additional problems beyond those in the field, constructing a larger dataset for geometric problem-solving. This dataset comprises a total of 6981 manually annotated and solved geometric problems.

    \item We established a geometric formal language system (FormalGeo) that guides the transformation of natural language into formal language. Concurrently, we built a geometric problem solver (FGPS) to validate annotated theorem sequences. By annotating the original image and text information of problems into a form understandable by the geometric formal language system, we can input the theorem sequences for validation. This not only ensures the correctness of proofs but also achieves interpretable step-by-step deduction.

    \item We introduced reinforcement learning, specifically employing Monte Carlo Tree Search, into geometric automated reasoning. This is done by conducting heuristic searches in a geometric reasoning environment built upon the FormalGeo system to address the limited annotated data issue.
\end{enumerate}

\section{Related Work}
\subsubsection{Datasets for Geometry Problems Solving}In the early research on geometric theorem proving\cite{early work1, early work2, early work3}, efforts such as those based on symbolic systems like GeoS\cite{Seo2015} or Inter-GPS aimed to construct formal language systems. These systems, by building solvers, enabled formal language to prove geometric propositions through logical deduction. However, with the rise of large language models, more recent work, including GeoQA and UniGeo, began treating problems and diagrams as multimodal inputs, generating solution sequences through large models. Although these methods exhibit high generality, they fall short in terms of interpretability and verifiability compared to symbol-based reasoning in formal systems.Past formal systems, designed manually for logical languages, faced limitations in broader application scenarios, resulting in relatively small datasets, such as GeoS with 186 questions and Geometry3k dataset which used by Inter-GPS containing approximately 3002 problems. These datasets appear relatively small when considered in the context of large models relying on extensive corpora. To address this issue, we integrated multiple publicly available geometric problem datasets. Due to the inconsistency in their formats, we manually reannotated them, including Geometry3k, GeoQA, and GeoQA+. Ultimately, we introduced a large-scale public dataset named FormalGeo7k, comprising around 6981 planar geometry problems.Our work is dedicated to addressing the heterogeneity in data sources to create a more unified and comprehensive dataset. This integration process aims to enhance data consistency and comparability, providing a more robust and comprehensive foundation for our research. This approach allows us to more effectively explore and analyze key features and patterns in geometric data.

\subsubsection{Reinforcement Learning for Geometry Problems Solving}In recent years, numerous studies have demonstrated the application of reinforcement learning in the reasoning process. In ~\cite{theorem proving}, Monte Carlo simulation was employed for mathematical theorem proving, while deep reinforcement learning was utilized for interactive theorem proving ~\cite{Tacticzero} and solving Math word problems ~\cite{AMWPG, Mathdqn}. However, there has been limited exploration of applying reinforcement learning to geometric theorem proving. The most recent work involves using deep reinforcement learning to address geometric problems ~\cite{GeoDRL}. In our study, considering the challenges in defining a reward function, we adopt the Monte Carlo method for interactive search. We replace the immediate rewards obtained through the reward function with delayed rewards acquired through the completion of a full round of interactions. The reward is determined by the success of theorem solving in the Monte Carlo Tree Search process\cite{MCTS}, akin to the approach used in AlphaGo\cite{AlphaGo}. This modification allows us to use reward feedback to shorten the length of theorem sequences, ultimately achieving an optimal sequence.

\section{Geometry Formal System}
In the process of solving geometric problems, mathematical formal systems play a crucial role, facilitating the transformation from information accessible to humans to the formal language comprehensible by machines. Particularly, when solving problems using artificial intelligence, formal systems become pivotal. Simultaneously, the combination of manual annotation and formal systems ensures consistency between the description of geometric problems and the dataset format, eliminating confounding factors for assessing the performance of artificial intelligence models. However, the answers generated by artificial intelligence models need reliable validation. Considering that traditional manual verification methods are both inefficient and error-prone, we have established a geometric problem solver on the basis of FormalGeo to verify answers\cite{FormalGeo}.

\begin{figure}
    \centering
    \includegraphics[width=1\linewidth]{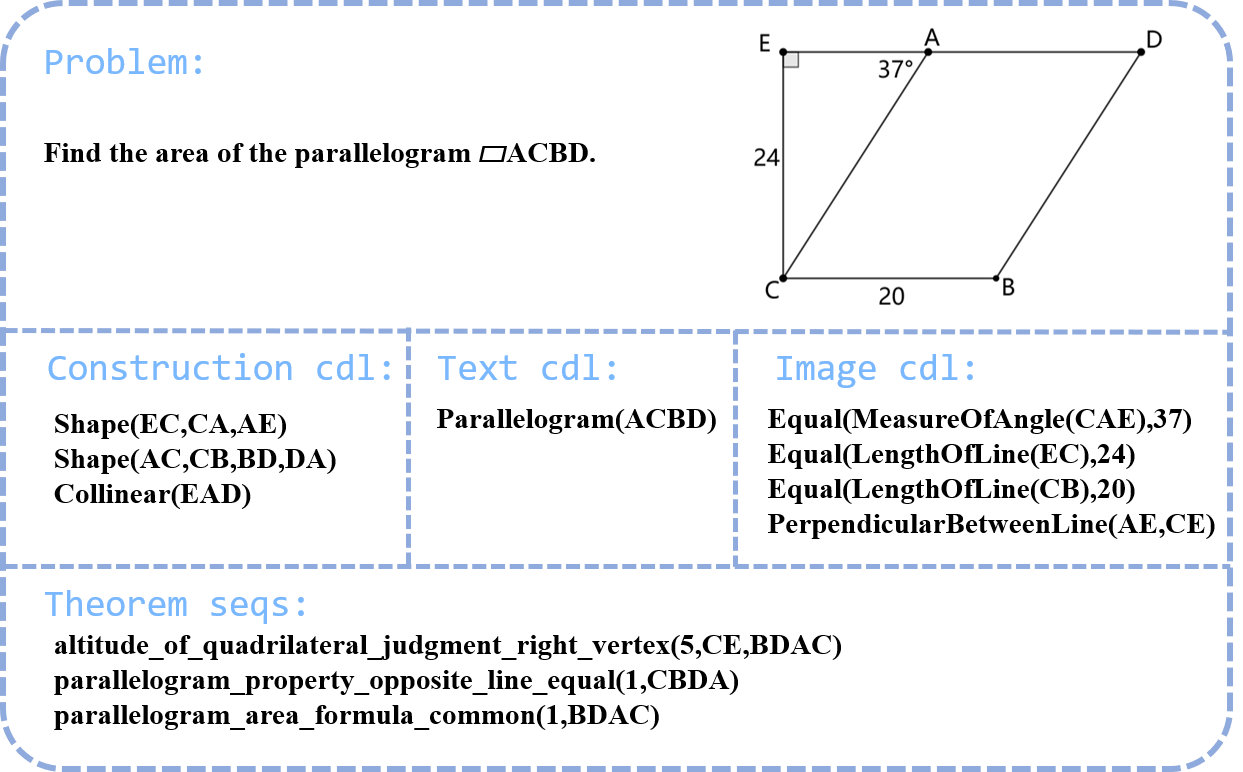}
    \caption{Formal Language for Geometry Problems}
    \label{fig:Problem}
\end{figure}

\subsection{Geometry Formal Language}The incorporation of a geometric formal system empowers computers with the capability to comprehend geometric problems, with formal language serving as the bridge between humans and machines. Within the geometric formal system, the formal language consists of Geometric Definition Language (GDL) and Conditional Declaration Language (CDL). As illustrated in Figure 1, CDL is utilized for describing textual and graphical information. GDL, on the other hand, is composed of Predicate Definition Language and Theorem Definition Language. Thus, the formal language functions as a bridge between humans and machines.

\textbf{Predicate Definition Language:} The Predicate Definition Language is employed to describe the direct geometric relationships and attributes among entities. It encompasses predicates that describe the structural aspects of geometric figures. Through structural predicates, one can articulate general relationships, such as coplanarity, shape, and collinearity. These structural predicates can further extend to basic entities like lines, angles, arcs, polygons, circles, and more. In addition to structural predicates, custom predicates can be utilized to automatically expand into other customized relationships. Examples include geometric relationships of custom shapes like right-angled triangles, trapezoids, as well as algebraic relationships.

\textbf{Theorem Definition Language:} It is comprised of premises and conclusions, with geometric relationships and attributes serving as premises, and conclusions being extended under the judgment rules of the theorem.

In the geometric formal system, FormalGeo, that we have constructed, it includes 88 predicates and 196 theorems. This system can represent, validate, and solve geometric problems ranging from SAT to IMO levels.

Ultimately, we can model the theorem-solving sequence as a hyper-tree. By continuously applying theorems to evaluate and integrate nodes in the tree, new nodes are generated. If the generated nodes align with the final state, it indicates successful problem-solving\cite{MDP}.

\subsection{FormalGeo7k Dataset}
Through the defined formal language, we integrated previously widely used public geometric datasets and online resources. Following formal rules, we annotated the datasets by declaring both graphical and textual information using the formal language. In addition to the problem information, we manually annotated theorem sequences based on the problem information, as illustrated in Figure \ref{fig:Problem}. These sequences were then validated using FGPS to ensure the correctness and consistency between the annotated theorem sequences and the declared problem information. Throughout the creation process of the FormalGeo7k dataset, we invested a total of 13 weeks, with the collaborative effort of 16 graduate students completing the annotation work. In total, we annotated 6981 problems.

The statistical data gathered from FormalGeo7k covers various aspects related to angle calculation, line length, perimeter, area, arc, and more. We categorize the difficulty levels of problems based on the lengths of theorem sequences, measured from L1 to L6. Here, L1 represents theorem sequence lengths within 2, L2 corresponds to lengths from 3 to 4, L3 encompasses lengths from 5 to 6, L4 covers lengths from 7 to 8, L5 includes lengths from 9 to 10, and L6 denotes lengths exceeding 10. Refer to Table \ref{tab:Datasets} for details.
\begin{table}[htbp]
\centering
\caption{Distribution of the FormalGeo7k Dataset}
\label{tab:Datasets}
\setlength{\tabcolsep}{3mm}
\begin{tabular}{cccccccc}
\toprule
\multirow{2}{*}{Category} & \multicolumn{7}{c}{Number} \\
\cmidrule{2-8}
 & Total & $L_1$ & $L_2$ & $L_3$ & $L_4$ & $L_5$ & $L_6$ \\
\midrule
    Total & 6981 & 2407 & 1898 & 1247 & 824 & 313 & 292\\
    Angle & 3523 & 1246 & 1160 & 535 & 378 & 127 & 77 \\
    Length & 2553 & 918 & 512 & 545 & 321 & 116 & 141\\
    Area & 366 & 106 & 96 & 67 & 44 & 17 & 36\\
    Perimeter & 305 & 22 & 81 & 74 & 56 & 41 & 31 \\
    Other & 234 & 115 & 49 & 26 & 25 & 12 & 7\\
\bottomrule
\end{tabular}
\end{table}

\section{Reasoner}
\subsection{Environment}
Our environment is modeled using a formalized system. In this formalized system, the process of applying theorems is abstracted into the generation of a hyper-tree. Each initial condition or condition derived through theorem deduction serves as a node. Multiple nodes are connected through the application of theorems, generating a new condition. If a node generated by a theorem is already the final answer needed, it is considered a successful problem-solving. The theorem process from the initial state to the terminating state can be represented using a theorem sequence.

In the environment utilized for reinforcement learning, as we need to utilize a natural language model for initial guidance, all initial conditions of the problems are considered as the initial state. Conditions obtained through the application of theorems are merged to form the next state for the reasoning process. By continuously applying theorems and ultimately reaching the goal state, the process is deemed successful.

\textbf{Reward:} In this environment, the rewards obtained from applying theorems cannot be quantified. Therefore, we set the reinforcement learning reward to be solely determined by the endpoint. In the context of geometric reasoning, regardless of its current state, if the problem is solvable, there exists a theorem sequence to answer it. Thus, when a theorem successfully resolves the problem, we assign a reward value of 1. For other cases, such as timeouts, non-executability of the theorem, or reaching the maximum step limit in the reasoning sequence, the reward is set to 0.

\begin{equation}
    Reward = \left\{
    \begin{array}{cc}
        1, & S_t \stackrel{a_t}{\Rightarrow} \textit{g} \\
        0, & \mathrm{otherwise}
    \end{array}
    \right.
\end{equation}

\textbf{Action Space:} We have set the number of theorems in our theorem library to be 196, and these theorems can be freely expanded as needed. Through the application of theorems, we can deduce new conditions based on several known conditions. However, due to strict requirements in constructing diagrams, multiple scenarios need to be considered. For instance, when determining the congruence of two triangles, if two sets of angles are known to be equal, it is sufficient for one set of three sides to be equal to establish the proof. Each side is rigorously defined in the diagram construction. Thus, on this basis, different branches of the theorem need to be used to prove the congruence relationship. Combining the 196 theorems with their branches, we can ultimately obtain 234 applicable theorems.

\subsection{Guidance}
For a mathematical problem under consideration, there are 234 executable action spaces from any given state. Therefore, if a geometric problem requires a lengthy theorem sequence, the search space explored by the search process will grow exponentially. Additionally, more challenging problems often involve more complex diagram constructions, leading to an increase in the time required to execute the theorems for such intricate diagrams. This is clearly unacceptable for the search process. In the case of random search, it needs to explore extensively to approximate the true probability distribution of the problem. If the time spent on exploration is excessively high, it can have unpredictable effects on the experiments. Hence, there is a need to leverage empirical knowledge for theorem guidance.

We pre-trained a lightweight Transformer network model, DistilBERT, with the aim of achieving prediction performance close to that of BERT while minimizing prediction time\cite{DistilBERT, BERT}. To accomplish this prediction goal, we processed the dataset containing 6981 problems. By applying annotated theorem sequences in the environment, we obtained the state of each node. The states, along with the executed theorems, were saved to the experience pool as $Exp(s_t, a_t, G_t)$. We then utilized a policy network to learn and predict the theorem selection for the current state.

\begin{algorithm}[H]
\caption{Agent Trajectory Prediction}
\label{alg:Trajectory}
\begin{algorithmic}[1] 
\State $s \gets env.init()$

\If{\textbf{not} $env.solved()$}
    \State $action\_probs \gets $ \textbf{RL\_PN$(s)$}
    \State $env.cur\_node.edges \gets \textbf{Normalize}(action\_probs, env.legal\_actions)$

    \Repeat
        \State $a \gets \textbf{UCB}(env.cur\_node.edges)$ \Comment{\textbf{Selection}}
        \State $s^{'} \gets env.step(a)$
    \Until{$a$  \textbf{not in} $env.visited\_edges$}

    \State $env.visited\_edges \gets a $ \Comment{\textbf{Expand}} 
    
    $\omega(s^{'},a) = 0$
    \For($i=0$; $i<simulation\_num$; $i++$)    
        \While{step in simulation\_steps} \Comment{\textbf{Simulation}}
            \If{$s^{'} \to g$}
                \State $\omega(s^{'},a) += 1$
                \State \textbf{break}
            \EndIf
            \State $a^{'} \gets $ \textbf{Sample}(\textbf{SL\_PN$(s^{'})$}
            \State $s^{''} \gets env.step(a^{'})$
        \EndWhile
    \EndFor
    
    $\omega(s^{'},a)  = \omega(s^{'},a)/simulation\_num$

    \State $env.backward(\omega(s^{'},a))$ \Comment{\textbf{Backward}}
    
\EndIf
\end{algorithmic}
\end{algorithm}

\subsection{Monte Carlo Tree Search}
Due to the data volume falling significantly short of what is required for language model training, relying solely on the policy network for theorem application is insufficient. We need to employ the policy network within the search process, where its inclusion substantially reduces the search space. Through Monte Carlo Tree Search, data collected during the exploration phase can be utilized for training the policy network, with the primary aim being to use the cumulative rewards obtained from actual exploration to inform the learning of the reinforcement learning policy network. The reinforcement learning network we use is initialized with parameters consistent with those of the policy network. As our learning objective is the probability distribution of theorems, the reinforcement learning network employs policy gradient learning during the training phase\cite{Policy Gradient}. By leveraging the cumulative reward feedback $R(\tau)$ obtained from Monte Carlo Tree Search \cite{MCTS, AlphaGo}, we can further reinforce the probability of making correct theorem selections. Algorithm \ref{alg:Trajectory} demonstrates how Monte Carlo Tree Search performs trajectory prediction. Our goal is to find parameters theta that maximize the objective function $R(\tau)$, that is, to find a policy that maximizes $E(R(\tau))$, where $\tau$ represents a trajectory from the initial state to the terminal state, and $R(\tau)$ is the cumulative reward garnered along this trajectory. In order to prevent overfitting of the policy network, we utilize the policy network obtained through supervised learning as our theorem predictor during the simulation phase. The parameters of this predictor remain consistent with the previous generation of reinforcement learning policy networks. In the theorem selection and expansion stages, we employ the latest reinforcement learning strategy for theorem selection\cite{DDQN}.

\begin{equation}
    J(\pi_\theta) = \underset{\theta}{\max}E_{\tau\sim\pi_\theta}\left[R(\tau)\right] 
\end{equation}

\begin{figure}
    \centering
    \includegraphics[width=1\linewidth]{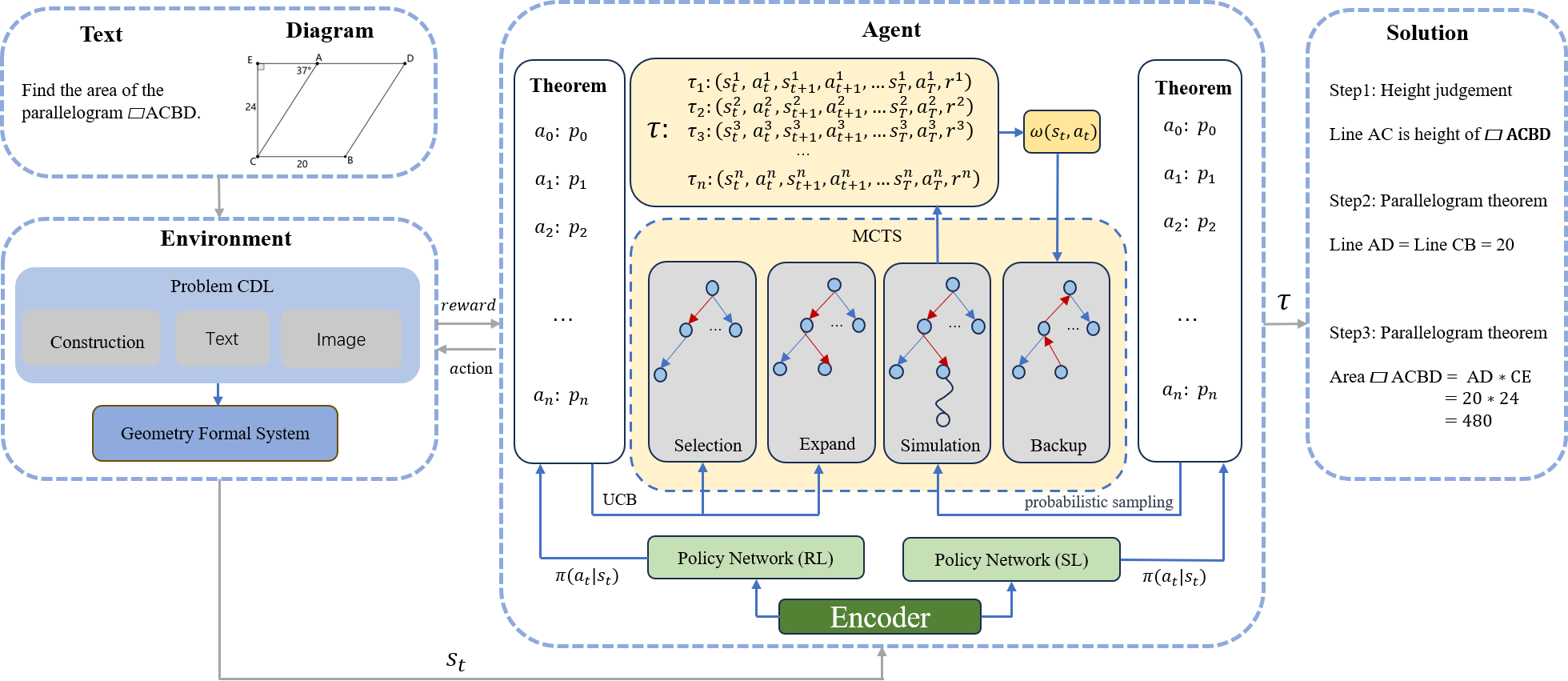}
    \caption{The overall architecture of the FGeoDRL model framework.}
    \label{fig:model}
\end{figure}

\noindent To maximize the cumulative reward, we need to optimize the path policy to enable it to select the strategy with the maximum cumulative reward. We use policy gradient to update the policy parameters:

\begin{equation}
    \nabla_\theta J(\pi_\theta) = E_{\tau\sim\pi_\theta}\left[\sum_{t=0}^T \nabla_\theta log\pi_\theta(a_t|s_t)R(\tau)\right]
\end{equation}

\begin{equation}
    \theta_{k+1} = \theta_k + \alpha\nabla_\theta J(\pi_\theta)
\end{equation}

To implement Monte Carlo Tree Search, we need to define relevant nodes and edges. Node S is described as composed of all the condition information under the current state, while an edge is described as an executable theorem in the action space. The node information includes the number of visits $N(s, a)$ for each node-action pair, the prior probability $P(s, a)$, and the  cumulative reward $G_t$ obtained by executing the corresponding action at that node\cite{Gt}. In the Monte Carlo Tree Search, we can divide it into the following four stages:

\textbf{Selection:} From the root node, we need to choose an optimal action $a_t$ to execute, and the selection of the optimal action is determined by the Upper Confidence Bound (UCB)\cite{UCB}. One part of UCB depends on the average reward obtained from exploration guided by the prior probability, i.e., the average reward obtained from exploration guided by the prior probability. The other part depends on the actual number of explorations selected during exploration. This is mainly to make the exploration more comprehensive, avoiding the drawback of continuously executing actions with high prior probabilities.

\begin{equation}
    a_t = \arg \underset{a}{\max}( G_t + c\sqrt{\frac{\ln{\sum_{i=0}^T}N(s_t, a_i)}{N(s_t, a_t)}})
\end{equation}

\textbf{Expansion:} For the action selected during the selection phase, the state node resulting from its execution is expanded. This expansion involves using the policy network to predict selectable theorems and leveraging prior probabilities to provide selection preferences. Subsequently, an unexpanded action $at$ is added to the tree, creating a new node $s_{t+1}$ by executing action $a_t$ . The goal of this phase is to further explore the selected node by simulating the outcome of the game and initialize statistical information for the newly expanded node.

\textbf{Simulation:} During the simulation phase, it is necessary to simulate the expanded state node. We set a maximum search step and simulation count, guided by the policy network to obtain more reliable simulation results within a smaller search space. For the simulation results, if the target node is reached within the maximum search steps, a reward of 1 is returned; otherwise, a reward of 0 is returned. Through multiple simulation processes, we obtain the average score $G_t$ for the expanded node.

\textbf{Backup:} For the reward scores and visit counts obtained during the simulation phase, we need to conduct backpropagation. Through backpropagation, we feed the simulation results $\omega(s_t, a_t)$ back to the search tree to update the cumulative rewards $G_t$, thereby influencing future theorem selections.

\begin{equation}
    G_t= R(\tau) = \sum_{t=0}^T\gamma^t r_{t} = \omega(s_t, a_t) + \gamma G_{t+1} 
\end{equation}

\section{Experiments}
\subsection{Train Method}
\subsubsection{Policy Network Datasets}In the initial phase of the project, we need to train a policy network for guiding theorem prediction. This involves processing an annotated dataset containing 6981 geometry problems. We partition this dataset into state-action pairs $(s_t, a_t)$ and delve into annotated theorems through a reinforcement learning environment. By referring to the ground truth answers, we can capture the states $s_t$ obtained at each step of theorem execution and the specific theorem applied $a_t$. This process not only helps establish the mapping between problems and actions but also provides crucial information for subsequent learning and optimization. Finally, we categorized the 6981 questions based on question types and difficulty levels. We divided them into training, testing, and validation sets in a ratio of 0.7:0.15:0.15, resulting in 20,981, 4,470, and 4,470 state-action pairs for each set, respectively. The experimental results on the validation set are presented in Table \ref{tab:PolicyNetwork}.

In Table \ref{tab:PolicyNetwork}, a range of 1 represents the direct hit rate of the prediction results, while values greater than 1 indicate the probability of the true label falling within the range space. As the training of the policy network is used to compress the search space, we expect the hit rate to indicate whether the label action exists within the predicted range. We sort the prediction results in the test set by probability and take the specified range of results as the search space. If the true label falls within the search space, it is considered a true positive. This to some extent reflects the extent to which the policy network compresses the search space, i.e., from the 234 theorems available in any state, compressing them into as few as 20 action spaces.

\begin{table}[H]
    \centering
    \caption{The hit rate of the policy network}
    \setlength{\tabcolsep}{3mm}
    \begin{tabular}{cccccccc}
    \toprule
      Range & 1 & 3 & 5 & 10 & 15 & 20 & 25  \\
    \midrule
        Hit Rate(\%)  & 49.35 & 72.89 & 81.74 & 89.67 & 93.21 & 94.76 & 95.91 \\
    \bottomrule
    \end{tabular}
    \label{tab:PolicyNetwork}
\end{table}

\subsubsection{Baseline}In the formalized system, we implemented several search baselines using different search strategies\cite{FormalGeo}. We compared forward search (FW) and backward search (BW) to evaluate distinct search policies. In the BW approach, the target node serves as the initial node, and the search progresses by continuously generating subgoals and solving them to achieve the overall search objective. FW and BW employ four main strategies, including breadth-first search (BFS), which searches layer by layer from the initial node. Depth-first search (DFS) conducts recursive depth-first traversal. Random search (RS) explores by randomly selecting actions at nodes. Beam search (BS) selects k actions for execution in the node expansion phase.

For the policy network trained through supervised learning, we will utilize it as the policy network for reinforcement learning through policy gradient learning. To achieve this, we conduct Monte Carlo Tree Search on the FormalGeo7k dataset. We set the simulation phase to 30 simulations with a maximum simulation length of approximately 30 steps. While the longest theorem sequence in the FormalGeo7k dataset reaches 25 steps, the majority of theorem sequences are within 10 steps. As the execution of theorems progresses, the state conditions become increasingly complex, leading to a significant increase in the execution time of the formalized system. If the simulation step length is set too long, it may result in a substantial increase in simulation time, severely impacting the efficiency of the search. Therefore, specific parameters need to be set for searching theorem sequences with step lengths exceeding 10. Refer to Table \ref{tab:FGeoDRL} for detailed results.

\begin{table}[H]
\centering
\caption{Comparison of search accuracy between FGeoDRL and the baseline.}
\label{tab:FGeoDRL}
\setlength{\tabcolsep}{3mm}
\begin{tabular}{cccccccc}
\hline
\multirow{2}{*}{Method} &  \multicolumn{7}{c}{Success Rates (\%)} \\
\cline{2-8}
& Total & $L_1$ & $L_2$ & $L_3$ & $L_4$ & $L_5$ & $L_6$ \\
\hline
FW-BFS & 38.86 & 59.95 & 38.62 & 28.55 & \textbf{17.35} & \textbf{8.63} & 3.77 \\
FW-DFS & 36.16 & 55.75 & \textbf{40.04} & 22.94 & 12.38 & 7.03 & 4.11 \\
FW-RS & \textbf{39.71} & 59.24 & \textbf{40.04} & \textbf{33.68} & 16.38 & 5.43 & \textbf{4.79} \\
FW-BS & 25.28 & 46.12 & 22.60 & 13.47 & 5.83 & 2.88 & 0.34 \\
BW-BFS & 35.44 & \textbf{67.22} & 33.72 & 11.15 & 6.67 & 6.07 & 1.03 \\
BW-DFS & 33.73 & 65.93 & 30.82 & 8.90 & 6.55 & 5.11 & 0.68 \\
BW-RS & 34.05 & 66.64 & 31.66 & 8.66 & 5.83 & 4.47 & 0.68 \\
BW-BS & 34.39 & 67.10 & 31.35 & 9.46 & 6.31 & 5.75 & 1.03 \\
FGeoDRL & 86.40 & 97.65 & 94.21 & 85.87 & 70.45 & 46.81 & 32.18\\
\hline
\end{tabular}
\end{table}

\subsection{Ablation Study}
In this section, we conduct ablation experiments on the FormalGeo7K dataset, considering three scenarios to explore the impact of different approaches on theorem prediction.

By directly predicting theorem sequences using the policy network, we set the number of theorems applied in each instance to 1, aiming to explore the direct accuracy of the policy network. Additionally, we employ a simulated search to showcase the generalization ability of the policy network to states. At each node in the search tree, we execute the n most probable theorems as the initial states for the next node. This approach generates a substantial amount of irrelevant information to serve as a confounding factor. Consequently, it results in node states that differ from those learned by the policy network, to some extent, demonstrating the generalization performance of the policy network.

Assessing the efficiency gained by combining regular search with policy network pruning, where we perform beam search to simulate the impact of policy network pruning\cite{Beam Search}.

Evaluating the improvement brought by the combination of Monte Carlo tree search and the policy network in FGeoDRL.

\begin{figure}[H]
    \centering
    \includegraphics[width=\textwidth]{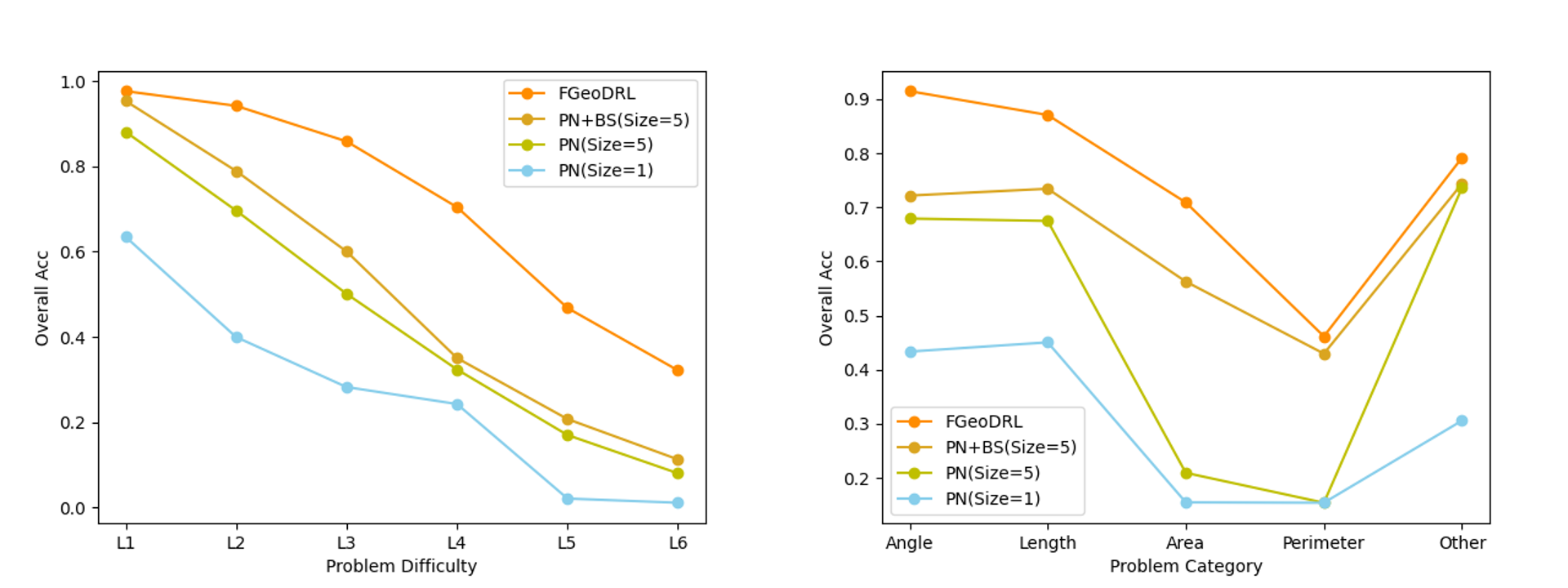}
    \caption{Performance of different strategies on the dataset}
    \label{fig:Distribution}
\end{figure}

The final experimental results are presented in Table ~\ref{tab:Ablation} and Figure ~\ref{fig:Distribution}.

Regarding the experimental outcomes, it is noticeable that the assistance of Monte Carlo Tree Search significantly boosts the problem-solving success rates from level $L_1$ to $L_3$. A considerable part of this success can be attributed to the sampling feedback during the simulation phase. For problems with shorter solution sequences, there is a higher probability to reach the solution endpoint within a finite number of steps. Upon reaching the endpoint, reward feedback facilitates the learning of problem-solving strategies by the policy network. This is particularly evident in dataset problems that involve area calculation, which, despite their lower quantity, tend to have shorter solution sequences and thus exhibit noticeably higher resolution rates compared to similar quantity problems with longer sequences, such as perimeter calculations. For problems with longer solution sequences, the performance relies more heavily on the policy network within the limited time. For problems with a greater number of questions regarding angles and lengths, their numerical advantage allows for an effective transition from short to long sequences, resulting in marked improvements in these two categories.

\begin{table}[H]
\centering
\caption{Ablation Study Results of FormalGeo7k.}
\label{tab:Ablation}
\setlength{\tabcolsep}{3mm}
\begin{tabular}{ccccc}
\hline
\multirow{2}{*}{Method} & \multirow{2}{*}{Size} & \multicolumn{3}{c}{Success Rates (\%)} \\
\cline{3-5}
& & Overall Acc(\%) & Avg Time(s) & Avg Steps\\
\hline
PN & 1 & 40.84 & 0.67 & 2.35\\
PN & 3 & 55.15 & 3.05 & 3.57\\
PN & 5 & 63.22 & 3.59 & 3.75\\
PN+BS & 3 & 61.79 & 3.41 & 3.26 \\
PN+BS & 5 & 70.61 & 3.57 & 3.43\\
FGeoDRL & / & 86.40 & / & 3.12\\
\hline
\end{tabular}
\end{table}

\section{Conclusion}
This paper proposes a theorem sequence prediction model, FGeoDRL, based on a reinforcement learning framework. We implemented a pre-trained natural language model for search pruning. Additionally, we interacted Monte Carlo tree search with a reinforcement learning environment built through symbolic reasoning systems. By conducting searches on our proposed geometric problem dataset, FormalGeo7k, we validated the deductive reasoning of geometric problems. FGeoDRL demonstrates superior interpretability, utilizing guided search and interactive feedback to correct experience networks. It plays a positive role in exploring both the shortest and novel solutions to geometric problems.

\section*{Acknowledgement}

This research was supported by NSFC Grant.12071282.

\end{document}